\documentclass[sigconf]{acmart}
\usepackage{booktabs} % For formal tables
\usepackage[utf8]{inputenc}
\usepackage{epsfig}
\usepackage{graphicx}
\usepackage{xcolor}
\usepackage{subcaption}
\usepackage{caption}
\usepackage{amsmath}
\usepackage{hyperref}
\usepackage{adjustbox}
\usepackage{multirow}
\usepackage{tabularx}
\usepackage[autostyle]{csquotes}
\usepackage{url}
\usepackage{algorithm}
\usepackage{algpseudocode}

\DeclareMathOperator*{\argmax}{arg\,max}

\AtBeginDocument{%
  \providecommand\BibTeX{{%
    \normalfont B\kern-0.5em{\scshape i\kern-0.25em b}\kern-0.8em\TeX}}}

\copyrightyear{2022}
\acmYear{2022}
\setcopyright{acmcopyright}\acmConference[WWW '22]{Proceedings of the ACM Web Conference 2022}{April 25--29, 2022}{Virtual Event, Lyon, France}
\acmBooktitle{Proceedings of the ACM Web Conference 2022 (WWW '22), April 25--29, 2022, Virtual Event, Lyon, France}
\acmPrice{15.00}
\acmDOI{10.1145/3485447.3512032}
\acmISBN{978-1-4503-9096-5/22/04}

\begin{document}
\sloppy\hyphenpenalty=3500
%%
%% The "title" command has an optional parameter,
%% allowing the author to define a "short title" to be used in page headers.
\title{WebFormer: The Web-page Transformer for Structure Information Extraction}
\author{Qifan Wang}
\authornote{Corresponding Authors.}
\affiliation{%
  \institution{Facebook AI}
  \city{Menlo Park}
  \state{CA}
  \country{USA}}
\email{wqfcr@fb.com}

\author{Yi Fang}
\affiliation{%
  \institution{Santa Clara University}
  \city{Santa Clara}
  \state{CA}
  \country{USA}}
\email{yfang@scu.edu}

\author{Anirudh	Ravula}
\affiliation{%
  \institution{Google Research}
  \city{Mountain View}
  \state{CA}
  \country{USA}}
\email{ravulaanirudh25@gmail.com}

\author{Fuli Feng}
\affiliation{%
 \institution{University of Science and Technology of China}
 \city{Hefei}
 \country{China}}
\email{fulifeng93@gmail.com}

\author{Xiaojun	Quan}
\affiliation{%
  \institution{Sun Yat-sen University}
  \city{Guangzhou}
  \country{China}}
\email{quanxj3@mail.sysu.edu.cn}

\author{Dongfang Liu}
\authornotemark[1]
\affiliation{%
  \institution{Rochester Institute of Technology}
  \city{Rochester}
  \state{NY}
  \country{USA}}
\email{dongfang.liu@rit.edu}
%%
%% The abstract is a short summary of the work to be presented in the
%% article.
\begin{abstract}
Structure information extraction refers to the task of extracting structured text fields from web pages, such as extracting a product offer from a shopping page including product title, description, brand and price. It is an important research topic which has been widely studied in document understanding and web search. Recent natural language models with sequence modeling have demonstrated state-of-the-art performance on web information extraction. However, effectively serializing tokens from unstructured web pages is challenging in practice due to a variety of web layout patterns. Limited work has focused on modeling the web layout for extracting the text fields.
In this paper, we introduce WebFormer, a Web-page transFormer model for structure information extraction from web documents. First, we design HTML tokens for each DOM node in the HTML by embedding representations from their neighboring tokens through graph attention. Second, we construct rich attention patterns between HTML tokens and text tokens, which leverages the web layout for effective attention weight computation.
%WebFormer therefore explicitly recovers both local syntactic and global layout information that may have been lost during serialization.
We conduct an extensive set of experiments on SWDE and Common Crawl benchmarks. Experimental results demonstrate the superior performance of the proposed approach over several state-of-the-art methods.
\end{abstract}

%%
%% The code below is generated by the tool at http://dl.acm.org/ccs.cfm.
%% Please copy and paste the code instead of the example below.
%%
\begin{CCSXML}
<ccs2012>
<concept>
<concept_id>10010147.10010178.10010179.10003352</concept_id>
<concept_desc>Computing methodologies~Information extraction</concept_desc>
<concept_significance>500</concept_significance>
</concept>
</ccs2012>
\end{CCSXML}

\ccsdesc[500]{Computing methodologies~Information extraction}

%%
%% Keywords. The author(s) should pick words that accurately describe
%% the work being presented. Separate the keywords with commas.
\keywords{web page extraction, structure extraction, transformer}

%%
%% This command processes the author and affiliation and title
%% information and builds the first part of the formatted document.
\maketitle

\begin{figure}[th]
\begin{center}
\includegraphics[width=0.85\linewidth]{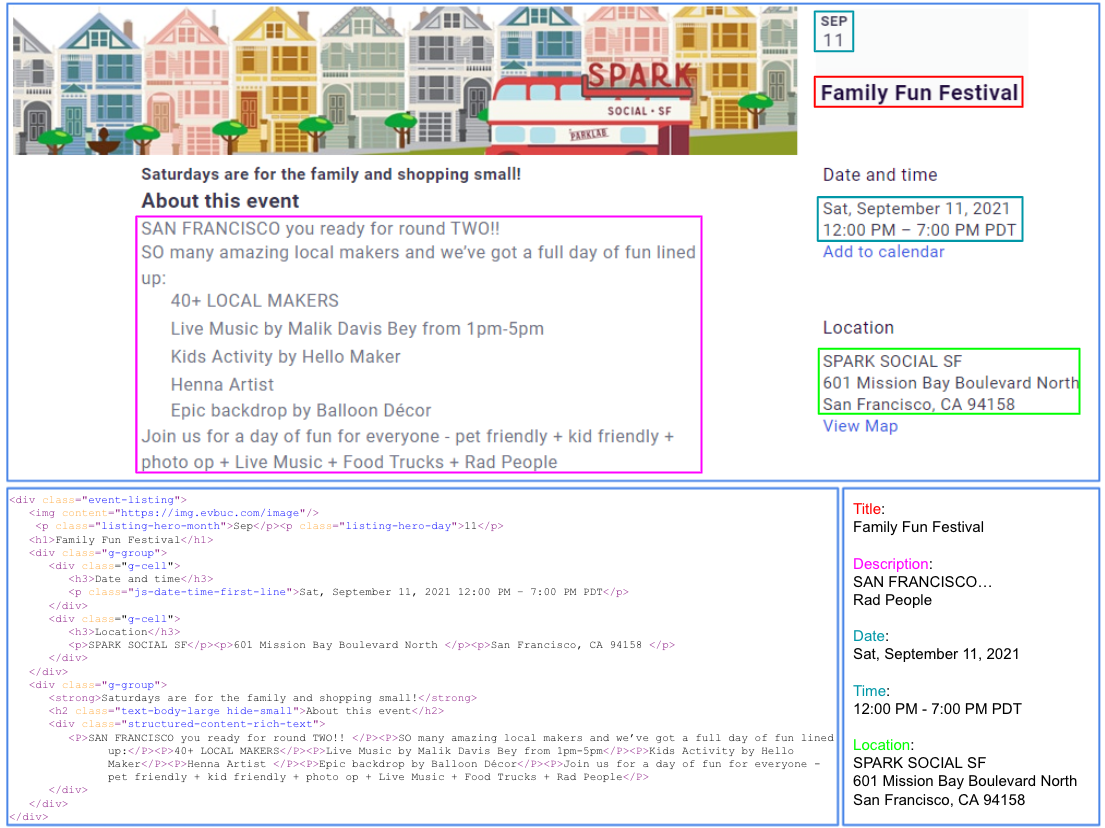}
\end{center}
\caption{An example of an event web page with its HTML (bottom left) and the extracted structured event information (bottom right), including event title, description, date and time, and location. The corresponding extractions of all text fields are highlighted with colored bounding boxes.} \label{fig:example}
\end{figure}
\section{Introduction}
%what is structure information extraction, and its applications, why it is important.
Web pages or documents are the most common and powerful source for humans to acquire knowledge.
There are billions of websites that contain rich information about various objects. For example, Figure \ref{fig:example} shows a web page describing an event, which contains structured event information including event title, description, date, time and location.
The large-scale web data becomes increasingly essential to facilitate new experiences in applications like web search and retrieval, which enables smart assistants to do complex tasks such as ``locating kid-friendly events in San Francisco this weekend'' and ``exploring Nike running shoes less than $\$$50''.
Therefore, it is an important research problem to extract structured information from web pages. %In this work, the problem of structure information extraction is to extract all the text spans for a set of target fields from a web document.

%existing methods - deep neural network to sequence modeling - what are the challenges - example figure
Structure information extraction from the web ~\cite{CrescenziM04,CarlsonS08,HaoCPZ11,ManabeT15,ZhanZ20} is a challenging task due to the unstructured nature of textual data and the diverse layout patterns of the web documents ~\cite{SleimanC13,MajumderPTWZN20}. There has been a lot of interest in this topic, and a plethora of research ~\cite{Yang00BN20,ChengQSH020,ZhangYXHLLLS21,TangXJWCXWWL21} in this area both in academia and industry.
Among the early works, template/wrapper induction ~\cite{DalviKS11,ProskurniaCPKWK17,LockardDSE18} has proven to be successful for extracting information from web documents. However, these techniques do not scale to the whole web as obtaining accurate ground truth for all domains is expensive. Moreover, the wrappers go out-of-date quickly because page structure changes frequently, and require periodic updating. One also needs to generate new templates for the new domains.

Recently, learning-based models ~\cite{GogarHS16,WangKGS19} have been proposed for automatic information extraction. These methods use schema.org markup \cite{TempelmeierDD18} as the supervision to build machine-learned extractors for different fields. Most recently, with the advance of natural language processing ~\cite{VaswaniSPUJGKP17,DevlinCLT19,abs-2004-05150}, language models with sequence modeling ~\cite{abs-2107-06955,WangWTJMDH21} have been applied to web document information extraction. These approaches first sequentialize the web document into a sequence of words, and then use RNN/LSTM \cite{ZhengMD018,Lin0VT20,abs-2101-02415} or attention networks ~\cite{XuXL0WWLFZCZZ20,HwangYPYS21} to extract the text spans corresponding to the structured fields from the sequence. Although existing natural language models achieve promising results on web information extraction, there are several major limitations.
First, the structural HTML layout has not been fully exploited, which contains important information and relation about different text fields.
For example, in an event page, the event date and location are naturally correlated, which form sibling nodes in the HTML (see Figure \ref{fig:example}). In a shopping page, the product price is often mentioned right after the product title on the page. Therefore, encoding the structural HTML beyond sequential modeling is essential in web document extraction.
Second, most existing models do not scale up to a large number of fields across domains. They build one separate model for each text field, which are not suitable for large scale extraction, nor can be generalized to new domains.
Third, large web documents with long sequences are not modeled effectively. Attention networks, such as Transformer-based models, usually limit their input to 512 tokens due to the quadratic computational cost with the sequence length.

%what we propose, experiments, contribution
In this paper, we propose WebFormer, a novel Web-page transFormer model that incorporates the HTML layout into the representation of the web document for structure information extraction. WebFormer encodes the field, the HTML and the text sequence in a unified Transformer model. Specifically, we first introduce HTML tokens for each DOM node in the HTML. We then design rich attention patterns for embedding representation among all the tokens.
%, including HTML-to-HTML attention via graph attention network, Text-to-Text token with relative attention, HTML-to-Text attention and Text-to-HTML attention.
WebFormer leverages the web layout structure for more effective attention weight computation, and therefore explicitly recovers both local syntactic and global layout information of the web document.
We evaluate WebFormer on SWDE and Common Crawl benchmarks, which shows superior performance over several state-of-the-art methods. The experimental results also demonstrate the effectiveness of WebFormer in modeling long sequences for large web documents. Moreover, we show that WebFormer is able to extract information on new domains.
We summarize the main contributions as follows:
\begin{itemize}
\item We propose a novel WebFormer model for structure information extraction from web documents, which effectively integrates the web HTML layout via graph attention.
\item We introduce a rich attention mechanism for embedding representation among different types of tokens, which enables the model to encode long sequences efficiently. It also empowers the model for zero-shot extractions on new domains.
\item We conduct extensive experiments and demonstrate the effectiveness of the proposed approach over several state-of-the-art baselines.
\end{itemize}

%rest of the paper
% The rest of the paper is organized as follows. Section 2 reviews the related work.
% Section 3 formally defines our problem and presents the proposed approach. Experimental setting and result are discussed in Section 4. The last section provides conclusions and points out possible future directions.

\section{Related Work}
% Network embedding focuses on generating the low-dimensional vector representation of nodes for real networks or graphs to facilitate further analysis of networks. Traditional approaches to network embedding can be divided into two categories: single network methods and multiplex ones. We review methods in both categories in the following subsections. Moreover, we provide discussions over the existing methods on partial data learning.

\subsection{Information Extraction}
Early studies of extracting information from the web pages mainly focus on building templates for HTML DOM tree, named wrapper induction ~\cite{CohenHJ02,KimS11}. Template extraction techniques have been applied to improve the performance of search engines, clustering, and classification of web pages. They learn desired patterns from the unstructured web documents and construct templates for information extraction. Region extraction methods ~\cite{ChangKGS06,SleimanC13} try to classify portions of a web page according to their specific purposes, e.g., classify whether a text node is the title field. Foley et al. ~\cite{FoleyBJ15} use simple naive-Bayes to classify the web page and SVM methods to get the score for each field. Wang et al. ~\cite{WangKGS19} extend this work by designing deep neural network models and using well designed visual features like font sizes, element sizes, and positions.
\begin{figure*}
\begin{center}
\includegraphics[width=0.84\linewidth]{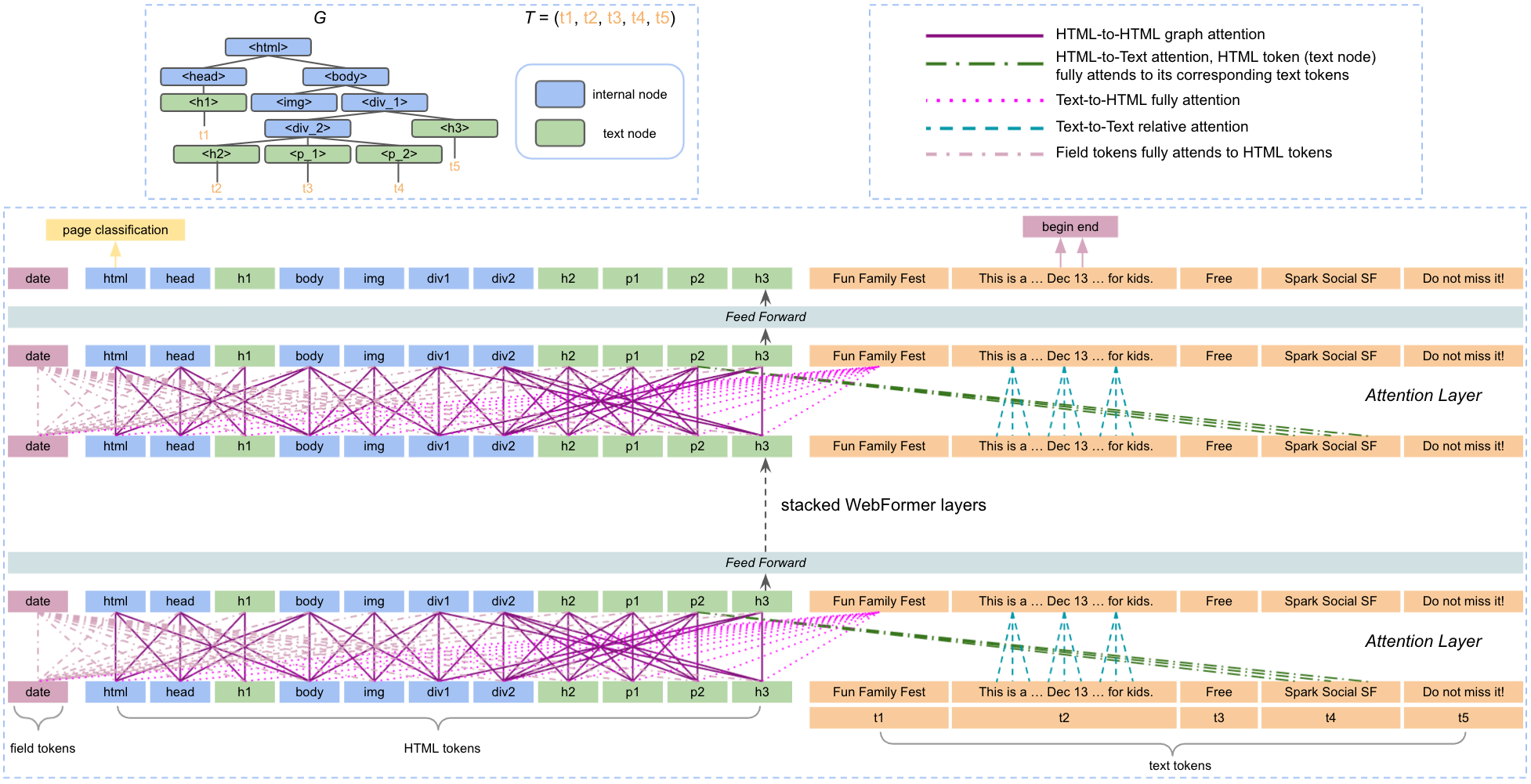}
\end{center}
\caption{The WebFormer model architecture.} \label{fig:overview}
\end{figure*}

%nlp models, multimodal nlp models
Recently, there has been an increasing number of works that develop natural language models with sequence modeling ~\cite{HuangXY15,MaH16,abs-2101-09465,Lin0VT20,abs-2102-09550,abs-2101-02415} for web information extraction.
Zheng et al. \cite{ZhengMD018} develop an end-to-end tagging model utilizing BiLSTM, CRF, and attention mechanism without any dictionary.
Aggarwal et al. \cite{AggarwalGSK20} propose a sequence-to-sequence model using an RNN, which leverages relative spatial arrangement of structures. Aghajanyan et al. \cite{abs-2107-06955} train a hyper-text language model based on BART \cite{BART} on a large-scale web crawl for various downstream tasks.
More recently, several attribute extraction approaches ~\cite{XuWMJL19,WangYKSSSYE20,Amazon2} have been proposed, which treat each field as an attribute of interest and extract its corresponding value from clean object context such as web title.
Chen et al. \cite{abs-2101-09465} formulate the web information extraction problem as structural reading comprehension and build a BERT \cite{DevlinCLT19} based model to extract structured fields from the web documents.
It is worth mentioning that there are also methods that work on multimodal information extraction ~\cite{YangYAKKG17,XuWLZM21,WangSLHDJ21,WangWTJMDH21}, which focus on extracting the field information from the visual layout or the rendered HTML of the web documents.
%However, most of these methods treat each attribute independently and build one separate model for each of them, which are not suitable for large scale attribute systems. Moreover, model generalization is not considered, which is important in zero-shot extraction.

\subsection{Relation Learning}
Relation extraction/learning research ~\cite{ZhengLWYZ16,HeCLZZZ18,LiLSZYHJ20,LiuCWZLX20,LockardSDH20,XuCZ21} is also related to our work.
Relation extraction refers to the task of extracting relational tuples and putting them in a knowledge base. Web information extraction can be thought of as the problem where the subject is known (the web document), and given the field (the relation) extract the corresponding text.
However, relation extraction has traditionally focused on extracting relations from sentences relying on entity linking systems to identify the subject/object and building models to learn the predicates in a sentence ~\cite{LevySCZ17,0005FC0S19}. Whereas in structure information extraction, usually the predicates (the fields) rarely occur in the web documents, and entity linking is very hard because the domain of all entities is unknown.

\section{WebFormer}
\subsection{Problem Definition}
We formally define the problem of structure information extraction from web documents in this section. The web document is first processed into a sequence of text nodes and the HTML DOM tree.
We denote the text sequence from the web document as $T = (t_1, t_2, \dots, t_k)$, where $t_i$ represents the $i$-$th$ text node on the web. $k$ is the total number of text nodes with $t_i$=$(w_{i_1},w_{i_2},\dots,w_{i_{n_i}})$ as its $n_i$ words/tokens. Note that the ordering of the text nodes does not matter in our model, and one can traverse the DOM tree in any order to obtain all the text nodes.
Denote the DOM tree of the HTML as $G$ = $(V, E)$, where $V$ is the set of DOM nodes in the tree with $E$ being the set of edges (see top left in Figure \ref{fig:overview}). Note that the $k$ text nodes are essentially connected in this DOM representation of the HTML, representing the layout of the web document.

The goal of structure information extraction is that given a set of target fields $F$=$(f_1,\dots,f_m)$, extract their corresponding text information from the web document. For example, for the text field ``date'', we aim to extract the text span ``Dec 13'' from the web document. Formally, the problem is defined as finding the best text span $\bar{s_j}$ for each field $f_j$, given the web document $T$ and $G$:
\[\bar{s_j} = \argmax_{b_j,e_j} \ Pr( \ w_{b_j}, \ w_{e_j} \ | \ f_j, \ T, \ G)\]
where $b_j$ and $e_j$ are the begin and end offsets of the extracted text span in the web document for text field $f_j$.

\subsection{Approach Overview}
Existing sequence modeling methods either directly model the text sequence from web document ~\cite{Lin0VT20,WangYKSSSYE20} or serialize the HTML with the text in a certain order ~\cite{abs-2101-09465,abs-2101-02415} to perform the span based text extraction. In this work, we propose to simultaneously encode the text sequence using the Transformer model and incorporate the HTML layout structure with graph attention.

The overall model architecture is shown in Figure \ref{fig:overview}. Essentially, our WebFormer model consists of three main components, the input layer, the WebFormer encoder and the output layer.
The input layer contains the construction of the input tokens of WebFormer as well as their embeddings, including the field token, the HTML tokens from DOM tree $G$ and the text tokens from the text sequence $T$.
The WebFormer encoder is the main block that encodes the input sequence with rich attention patterns, including HTML-to-HTML (H2H), HTML-to-Text (H2T), Text-to-HTML (T2H) and Text-to-Text (T2T) attentions.
In the output layer, the text span corresponding to the field is computed based on the encoded field-dependent text embeddings.
We present the detail of each component separately in the following subsections.

\subsection{Input Layer}
Most previous sequence modeling approaches ~\cite{AggarwalGSK20,Amazon2} only encode the text sequence of the web document without utilizing the HTML layout structure.
In this work, we jointly model the text sequence with the HTML layout in a unified Transformer model. In particular, we introduce three types of tokens in the input layer of WebFormer.

\noindent\textbf{Field token} A set of field tokens are used to represent the text field to be extracted, such as ``title'', ``company'' and ``base salary'' for a job page. By jointly encoding the text field, we are able to construct a unique model across all text fields.

\noindent\textbf{HTML token} Each node in the DOM tree $G$, including both internal nodes (non-text node) and text nodes, corresponds to an HTML token in WebFormer. The embedding of a HTML token can be viewed as a summarization of the sub-tree rooted by this node. For example, in Figure \ref{fig:overview}, the embedding of the ``$<$$html$$>$'' token essentially represents the full web document, which can be used for page level classification. On the other hand, the embedding of the text node ``$<$$p_2$$>$'' summarizes the text sequence $t_4$.

\noindent\textbf{Text token} This is the commonly used word representation in natural language models. For example, $t_1$ contains three words, ``Fun'', ``Family'' and ``Fest'', which correspond to three text tokens.

In the input layer, every token is converted into a $d$-dimensional embedding vector. Specifically, for field and text tokens, their final embeddings are achieved by concatenating a word embedding and a segment embedding. For HTML token embedding, they are formulated by concatenating a tag embedding and a segment embedding.
The word embedding is widely adopted in the literature \cite{MikolovSCCD13}.
The segment embedding is added to indicate which type the token belongs to, i.e. field, HTML or text.
The tag embedding is introduced to represent different HTML-tag of the DOM nodes, e.g. ``$div$'', ``$head$'', ``$h1$'', ``$p$'', etc.
Note that all the embeddings in our approach are trainable. The word embeddings are initialized from the pretrained language model, while the segment and tag embeddings are randomly initialized.

\subsection{WebFormer Encoder}
The WebFormer encoder is a stack of $L$ identical contextual layers, which efficiently connects the field, HTML and text tokens with rich attention patterns followed by a feed-forward network. The encoder produces effective contextual representations of web documents. To capture the complex HTML layout with the text sequence, we design four different attention patterns, including 1) HTML-to-HTML (H2H) attention which models the relations among HTML tokens via graph attentions. 2) HTML-to-Text (H2T) attention, which bridges the HTML token with its corresponding text tokens. 3) Text-to-HTML (T2H) attention that propagates the information from the HTML tokens to the text tokens. 4) Text-to-Text (T2T) attention with relative position representations. Moreover, WebFormer incorporates the field into the encoding layers to extract the text span for the field.

\subsubsection{HTML-to-HTML Attention}
%graph construction
The HTML tokens are naturally connected via the DOM tree graph. The H2H attention essentially computes the attention weights among the HTML tokens and transfers the knowledge from one node to another with the graph attention \cite{VelickovicCCRLB18}. We use the original graph $G$ that represents the DOM tree structure of the HTML in the H2H attention calculation. In addition, we add edges to connect the sibling nodes in the graph, which is equivalent to include certain neighbors with edge distance 2 in the graph. For example, the HTML token ``$<$$div1$$>$'' is connected with itself, the parent token ``$<$$body$$>$'', the child tokens ``$<$$div2$$>$'' and ``$<$$h3$$>$'', and sibling token ``$<$$img$$>$''. Formally, given the HTML token embedding $x_i^H$, the H2H graph attention is defined as:
\[\alpha^{H2H}_{ij} = \frac{\exp(e^{H2H}_{ij})}{\sum_{\ell\in \mathcal{N}(x_i^H)} \exp(e^{H2H}_{i\ell})}, \ for \ j \in \mathcal{N}(x_i^H)\]
\[e^{H2H}_{ij} = \frac{x^H_i W_Q^{H2H} (x^H_j W_K^{H2H} + a_{ij}^{H2H})^T}{\sqrt{d}}\]
\noindent where $\mathcal{N}(x_i^H)$ indicates the neighbors of the HTML token $x_i^H$ in the graph. $W_Q^{H2H}$ and $W_K^{H2H}$ are learnable weight matrices, and $a_{ij}^{H2H}$ are learnable vectors representing the edge type between the two nodes, i.e. parent, child or sibling. $d$ is the embedding dimension.

\subsubsection{HTML-to-Text Attention}
%only for text node
The H2T attention is only computed for the text nodes in the HTML to update their contextual embeddings.
We adopt a full attention pattern where the HTML token $x_i^H$ is able to attend to each of its text tokens $x_j^T$ in $t_i$.
For example, in Figure \ref{fig:overview}, the HTML token ``$<$$p_2$$>$'' attends to all the three text tokens in $t_4$, i.e. ``Spark'', ``Social'' and ``SF''. The H2T full attention is defined as:
\[\alpha^{H2T}_{ij} = \frac{\exp(e^{H2T}_{ij})}{\sum_{\ell\in t_i} \exp(e^{H2T}_{i\ell})}, \ for \ j \in t_i\]
\[e^{H2T}_{ij} = \frac{x^H_i W_Q^{H2T} (x^T_j W_K^{H2T})^T}{\sqrt{d}}\]
\noindent where $W_Q^{H2T}$ and $W_K^{H2T}$ are weight matrices in H2T attention.

\subsubsection{Text-to-HTML Attention}
%fully attention
In T2H attention, each text token communicates with every HTML token. Intuitively, this T2H attention allows the text token to absorb the high-level representation from these summarization tokens of the web document. The formulation of the T2H attention is analogous to the above H2T attention except that each text token attends to all HTML tokens.

\subsubsection{Text-to-Text Attention}
%relative attention
The T2T attention is the regular attention mechanism used in various previous models ~\cite{VaswaniSPUJGKP17,DevlinCLT19}, which learns contextual token embeddings for the text sequence.
However, the computational cost of the traditional full attention grows quadratically with the sequence length, and thus limits the size of the text tokens. Inspired by the work of ~\cite{ShawUV18,ShawMCPA19}, our T2T attention adopts relative attention pattern with relative position encodings, where each text token only attends to the text tokens within the same text sequence and within a local radius $r$. In Figure \ref{fig:overview}, the local radius $r$ is set to 1, which means each token will only attend to its left and right tokens, and itself. For instance, the text token ``is'' in $t_2$ attends to the tokens ``This'', ``is'' and ``a'' within $t_2$. The formal T2T relative attention is defined as:
\[\alpha^{T2T}_{ij} = \frac{\exp(e^{T2T}_{ij})}{\sum_{i-r\le\ell\le {i+r}} \exp(e^{T2T}_{i\ell})}, \ for \ i-r \le j \le i+r\]
\[e^{T2T}_{ij} = \frac{x^T_i W_Q^{T2T} (x^T_j W_K^{T2T} + b_{i-j}^{T2T})^T}{\sqrt{d}}\]
\noindent where $W_Q^{T2T}$ and $W_K^{T2T}$ are weight matrices in T2T attention. $b_{i-j}^{T2T}$ are learnable relative position encodings representing the relative position between the two text tokens. Note that there are total $2r+1$ possible relative position encodings, i.e. ${(i-j)} \in\{-r,\dots,-1,0,1,\dots,r\}$.

\subsubsection{Field Token Attention}
%fully attention to the HTML tokens
Our WebFormer model jointly encodes the field information such that the structured fields share the unique encoder. Following the work in ~\cite{XuWMJL19,WangYKSSSYE20}, we introduce the field tokens into WebFormer and enable full cross-attentions between field and HTML tokens. Note that one can easily add cross-attention between field and text tokens. We found empirically in our experiments that this does not improve the extraction quality. Although there is no direct interaction between field and text tokens, they are bridged together through the text-to-HTML and the HTML-field attentions.

\subsubsection{Overall Attention}
We compute the final token representation based on the above rich attention patterns among field, text and HTML tokens.
The output embeddings for field, text and HTML tokens $z_i^F, z_i^T, z_i^H$, are calculated as follows:
\[z_i^F = \sum_{j} \alpha^{F2H}_{ij} x_j^H W_V^F\]
\[z_i^T = \sum_{i-r\le j\le {i+r}} \alpha^{T2T}_{ij} x_j^T W_V^T + \sum_{k} \alpha^{T2H}_{ij} x_k^H W_V^H\]
\[z_i^H = \sum_{j \in \mathcal{N}(x_i^H)} \alpha^{H2H}_{ij} x_j^H W_V^H + \sum_{k\in t_i} \alpha^{H2T}_{ij} x_k^T W_V^T\]
where all the attention weights $\alpha_{ij}$ are described above. $W_V^F$, $W_V^T$ and $W_V^H$ are the learnable matrices to compute the values for field, text and HTML tokens respectively.

\subsection{Output Layer}
The output layer of WebFormer extracts the final text span for the field from the text tokens. We apply a softmax function on the output embeddings of the encoder to generate the probabilities for the begin and end indices:
\[P_b = softmax(W_bZ^T), \ P_e = softmax(W_eZ^T) \]
where $Z^T$ is the contextual embedding vectors of the input text sequence. $W_b$ and $W_e$ are two parameter matrices that project the embeddings to the output logits, for the begin and end respectively. Inspired by the work \cite{XLNet}, we further predict the end index based on the start index by concatenating the begin token embedding with every token embedding after it.

\subsection{Discussion}
This section provides discussion that connects WebFormer with previous methods as well as the limitations of our model. If we treat HTML tags as additional text tokens, and combine with the text into a single sequence without the H2H, H2T and T2H attentions, our model architecture degenerates to the sequence modeling approaches ~\cite{XuL0HW020,abs-2101-09465} that serialize the HTML layout. If we further trim the HTML from the sequence, our model is regressed to the sequence model \cite{WangYKSSSYE20} that only uses the text information. Moreover, if we also remove the text field from the input, our model degenerates to the sequence tagging method ~\cite{ZhengMD018,Lin0VT20}, which is not able to scale to a large set of target fields.

There are two scenarios where our model is not directly applicable. First, our model focuses on structure information extraction on single object pages, where each target field only has one text value. For a multi-object page, e.g. a multi-event page, there are different titles and dates corresponding to different events on the page, which could be extracted with methods like repeated patterns \cite{AdelfioS13,WangKGS19}. Second, there are applications that require to extract information from the rendered pages, where OCR and CNN \cite{XuL0HW020} techniques are used.

\section{Experiments}
\subsection{Datasets}
%The proposed approach is evaluated on two benchmarks.
\noindent\textbf{SWDE} ~\cite{HaoCPZ11,abs-2101-02415}: The Structured Web Data Extraction (SWDE) dataset is designed for structural reading comprehension and information extraction on the web. It consists of more than 124,000 web pages from 80 websites of 8 verticals including ``auto'', ``book'', ``camera'', ``job'', ``movie'', ``nbaplayer'', ``restaurant'' and 'university'. Each vertical consists of 10 websites and contains 3 to 5 target fields of interest.
We further split the data into train, dev and test sets with 99,248, 12,425 and 12,425 pages respectively.
\begin{table}
\begin{adjustbox}{width=1\columnwidth,center}
\begin{tabular}{c|c|c|c|c}
\hline
\multirow{2}{*}{ Data Splits} &	\multirow{2}{*}{SWDE} & \multicolumn{3}{c}{Common Crawl} \\
\cline{3-5}
& &Events & Products & Movies \\
\hline
Train  &99,248   &72,367 &105,642 &57,238   \\
Dev/Test    &12,425 & 9,046  &  13,205  &7,154\\
\hline
Training Time (10 epoch)    &4h 15m & 3h 46m  &  4h 22m  &3h 21m\\
%Test  & 321 & 9,046 & 13,205  &7,154\\
\hline
\end{tabular}
\end{adjustbox}
\caption{Statistics of the datasets with the training time.} \label{table:data}
\end{table}

\noindent\textbf{Common Crawl}\footnote{\url{http://commoncrawl.org/connect/blog/}}: The Common Crawl corpus is widely used in various web search, information extraction and other related tasks. Common Crawl contains more than 250 TiB of content from more than 3 billion web pages. In our experiments, we select web pages that have schema.org annotations\footnote{\url{https://schema.org/}} within the three domains - \textbf{Events}, \textbf{Products} and \textbf{Movies}. The schema.org annotations contain the website provided markup information about the object, which are used as our ground-truth labels. The fields are \{``Name'', ``Description'', ``Date'', ``Location''\}, \{``Name'', ``Description'', ``Brand'', ``Price'', ``Color''\} and \{``Name'', ``Description'', ``Genre'', ``Duration'', ``Director'', ``Actor'', ``Published Date''\} for event, product and movie pages respectively. We further filter these pages by restricting to English and single object pages. We downsample the web pages by allowing at most 2,000 pages per website to balance the data, as some websites might dominate, e.g., amazon.com. All datasets are then randomly split into train, dev and test sets with raito 8:1:1. The details are given in Table \ref{table:data}.
\begin{table*}
\begin{adjustbox}{width=1.9\columnwidth,center}
\begin{tabular}{c|cc|cc|cc|cc}
\hline
\multirow{3}{*}{Models} &	\multicolumn{2}{c|}{\multirow{2}{*}{\bf SWDE}} & \multicolumn{6}{c}{\bf Common Crawl} \\
\cline{4-9}
& &	&\multicolumn{2}{c|}{\bf Events} &\multicolumn{2}{c|}{\bf Products}&\multicolumn{2}{c}{\bf Movies}\\
\cline{2-9}
& EM & F1 &EM & F1 & EM & F1 & EM & F1 \\
\hline
OpenTag & 81.33 $\pm$ 0.22 & 86.54 $\pm$ 0.27 & 77.14 $\pm$ 0.26 & 83.71 $\pm$ 0.12& 72.57 $\pm$ 0.20 & 77.75 $\pm$ 0.19 & 80.36 $\pm$ 0.15 & 85.06 $\pm$ 0.18 \\
DNN & 80.53 $\pm$ 0.15 & 85.64 $\pm$ 0.26 & 78.43 $\pm$ 0.18 & 85.06 $\pm$ 0.21& 74.64 $\pm$ 0.27 & 78.56 $\pm$ 0.15 & 82.44 $\pm$ 0.23 & 86.65 $\pm$ 0.16 \\
AVEQA & 83.27 $\pm$ 0.32 & 88.75 $\pm$ 0.16 & 80.82 $\pm$ 0.21 & 86.47 $\pm$ 0.14& 74.85 $\pm$ 0.32 & 79.49 $\pm$ 0.28 & 83.87 $\pm$ 0.30 & 88.51 $\pm$ 0.19 \\
SimpDOM & 84.67 $\pm$ 0.23 & 90.35 $\pm$ 0.21 & 81.96 $\pm$ 0.24 & 86.33 $\pm$ 0.17& 75.12 $\pm$ 0.27 & 78.22 $\pm$ 0.21 & 82.59 $\pm$ 0.25 & 87.72 $\pm$ 0.18 \\
H-PLM & 83.42 $\pm$ 0.20 & 89.04 $\pm$ 0.18 & 82.65 $\pm$ 0.15 & 87.52 $\pm$ 0.17& 76.24 $\pm$ 0.17 & 81.13 $\pm$ 0.26 & 83.72 $\pm$ 0.26 & 89.34 $\pm$ 0.17 \\
\hline
WebFormer & {\bf 86.58 $\pm$ 0.16} & {\bf 92.46 $\pm$ 0.24} & {\bf 84.79 $\pm$ 0.24} & {\bf 89.33 $\pm$ 0.18}& {\bf 80.67 $\pm$ 0.20} & {\bf 83.37 $\pm$ 0.23} & {\bf 85.30 $\pm$ 0.19} & {\bf 90.41 $\pm$ 0.24} \\
\hline
\end{tabular}
\end{adjustbox}
\caption{Performance comparison on all datasets. Results are statistically significant with p-value $<$ 0.001.}\label{table:performance}
\end{table*}

\begin{table}
\begin{adjustbox}{width=0.88\columnwidth,center}
\begin{tabular}{c|cc|cc|cc}
\hline
\multirow{2}{*}{Fields} &	\multicolumn{2}{c|}{\bf Events} & \multicolumn{2}{c|}{\bf Products} & \multicolumn{2}{c}{\bf Movies} \\
\cline{2-7}
& EM & F1 &EM & F1 & EM & F1 \\
\hline
Name & 88.27 & 93.46 & 85.11 & 90.53 & 89.32 & 93.57  \\
Description & 81.62 & 85.50 & 77.94 & 81.46 & 82.71 & 88.19  \\
Date & 86.86 & 91.48 & - & - & - & -  \\
Location & 82.41 & 86.88 & - & - & - & -   \\
Brand & - & - & 84.23 & 85.63  & - & -  \\
Price  & - & - & 75.65 & 76.86  & - & -  \\
Color  & - & - & 80.42 & 82.35  & - & - \\
Genre  & - & -  & - & - & 89.49 & 92.67  \\
Duration  & - & -  & - & - & 83.74 & 88.35  \\
Director  & - & -  & - & - & 86.28 & 91.38  \\
Actor  & - & - & - & - & 80.16 & 87.44  \\
Publish Date & - & -  & - & - & 85.40 & 91.27  \\
\hline
\end{tabular}
\end{adjustbox}
\caption{Field level metrics of WebFormer.}\label{table:field_performance}
\end{table}

\subsection{Implementation Detail}
For data pre-processing, we use open-source LXML library\footnote{\url{https://lxml.de/}}
to process each page for obtaining the DOM tree structures. We then use in order traverse of the DOM tree to obtain the text nodes sequence.
We implemented our models using Tensorflow and Keras. Each model is trained on a 32 core TPU v3 configuration.
The word embedding is initialized with the pretrained BERT-base. The parameters used in WebFormer are 12 layers, 768 hidden size, 3072 hidden units (for FFN) and 64 local radius. The maximum text sequence length is set to 2048. The maximum number of HTML tokens are set to 256.
During training, we use the gradient descent algorithm with Adam optimizer. The initial learning rate is set to $3e^{-5}$. The batch size for each update is set as 64 and the model is trained for up to 10 epochs. The dropout probability for the attention layer is set to 0.1.

\subsection{Evaluation Metric}
We evaluate the performance of the WebFormer model with two standard evaluation metrics: \textbf{Exact Match} (EM) and \textbf{F1} from the package released in \cite{squad}.
Exact Match is used to evaluate whether a predicted span is completely the same as the ground truth. It will be challenging for those answers that are only part of the text. F1 measures the overlap of the extracted answer and the ground truth by splitting the answer span into tokens and compute F1 score on them. We repeat each experiment 10 times and report the metrics on the test sets based on the average over these runs.

\subsection{Baselines}
%We compare our model with the following baselines.
\noindent\textbf{OpenTag} \cite{ZhengMD018} uses a BiLSTM-Attention-CRF architecture with sequence tagging strategies. OpenTag does not encode the field and thus builds one model per field.

\noindent\textbf{DNN} \cite{WangKGS19} applies deep neural networks for information extraction. Text nodes in the HTML are treated as candidates, and are extracted with DNN classifiers.

\noindent\textbf{AVEQA} \cite{WangYKSSSYE20} formulates the problem as an attribute value extraction task, where each field is treated as an attribute. This model jointly encodes both the attribute and the document with a BERT \cite{DevlinCLT19} encoder.

\noindent\textbf{SimpDOM} \cite{abs-2101-02415} treats the problem as DOM tree node tagging task by extracting the features for each text node including XPath, and uses a LSTM to jointly encode with the text features.

\noindent\textbf{H-PLM} \cite{abs-2101-09465} sequentializes the HTML together with the text and builds a sequence model using the pre-training ELECTRA \cite{ClarkLLM20} as backbone.

The codes for OpenTag\footnote{\url{https://github.com/hackerxiaobai/OpenTag_2019}} and H-PLM\footnote{\url{https://github.com/X-LANCE/WebSRC-Baseline}} are publicly available. For our previous works DNN and AVEQA, we use the original codes for the papers. For SimpDOM, we re-implement their model using the parameters from the paper.
\begin{figure}
\centering
\includegraphics[width=0.9\linewidth]{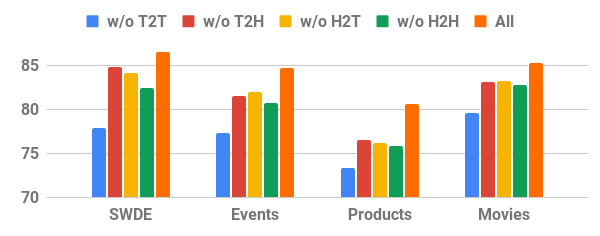}
\includegraphics[width=0.9\linewidth]{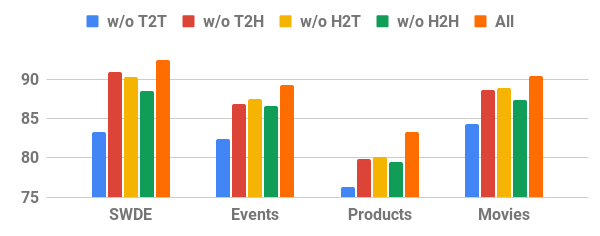}
\caption{Results of WebFormer with different attention patterns. Top: EM scores. Bottom: F1 scores.} \label{fig:att}
\end{figure}

\subsection{Results and Discussion}
\subsubsection{Performance Comparison}
The evaluation results of WebFormer and all baselines are reported in Table \ref{table:performance}. From these comparison results, we can see that WebFormer achieves the best performance among all compared methods on all datasets. For example, the EM metric of WebFormer increases over 7.8\% and 5.8\% compared with AVEQA and H-PLM on Products. There are three main reasons: First, our model integrates the HTML layout into a unified HTML-text encoder with rich attention, which enables the model to effectively understand the web layout structure. Second, WebFormer adopts the relative position encoding in T2T attention, which allows our model to represent large documents efficiently. Third, the field information is jointly encoded and attended with both HTML and text tokens. Different fields share one encoder and thus are able to benefit from each other.
We further report the field level results of WebFormer on the Common Crawl dataset in Table \ref{table:field_performance}. It can be seen that some fields, such as ``Name'' and ``Genre'', obtain relatively higher scores compared with ``Price'' and ``Location''. We also observe that the difference between EM and F1 scores is very small for fields like ``Brand'' and ``Color''. The reason is that their text spans are usually very short, containing just one or two tokens.
\begin{figure}
\centering
\includegraphics[width=0.95\linewidth]{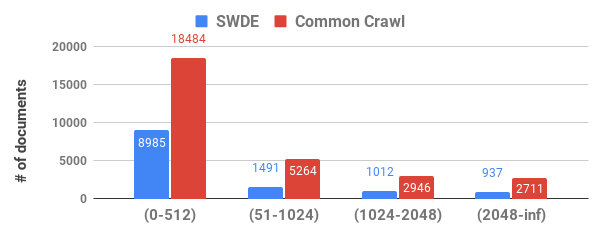}
\includegraphics[width=0.95\linewidth]{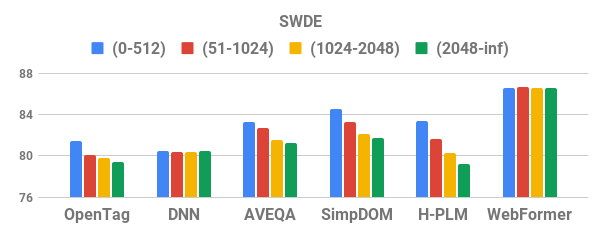}
\includegraphics[width=0.95\linewidth]{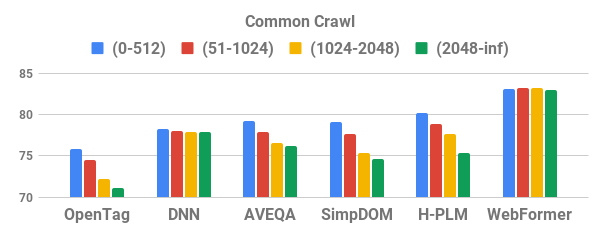}
\caption{EM scores of different methods within each bucket of sequence length.} \label{fig:len}
\end{figure}

\subsubsection{Impact of Rich Attentions}
To understand the impact of the rich attention patterns, we conduct a set of experiments by removing each attention from our model. Specifically, we train four separate models without T2T, H2T, T2H and H2H attention respectively.
The results of these four models and WebFormer (refer to All) on all datasets are shown in Figure \ref{fig:att}. It is not surprising to see that the performance drops significantly without the T2T local attention. The reason is that T2T is used to model the contextual token embeddings for the text sequence, which is the fundamental component in the Transformer model.
We can also observe that the model without H2H graph attention achieves much worse performance compared to the models without T2H or H2T attention. This observation validates that the HTML layout information encoded within the H2H attention is crucial for extracting structure fields from web documents. Moreover, it is clear that WebFormer with T2H and H2T attentions further improve the model performance on all datasets.
\begin{figure}
\centering
\includegraphics[width=0.92\linewidth]{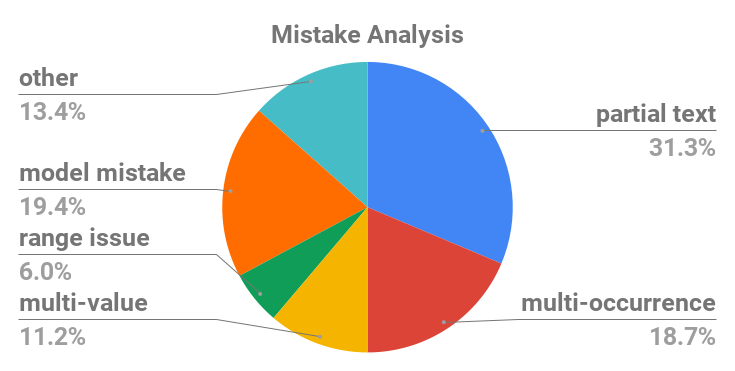}
\caption{Mistake analysis: distribution of different type of mistakes.} \label{fig:mistake}
\end{figure}

\subsubsection{Impact on Large Document}
To evaluate the impact of different models on large documents with long text sequence, we group the test examples into four buckets w.r.t. the sequence length of the example (i.e. 0-512, 512-1024, 1024-2048 and 2048-inf), and compute the metrics in each bucket for all methods. The length distribution of the test documents with EM scores on both datasets (for Common Crawl, we merge all the test sets from Events, Products and Movies) are shown in Figure~\ref{fig:len}. It can be seen that WebFormer achieves consistent results w.r.t. the sequence length. In contrast, the performances of OpenTag, AVEQA, SimpDOM and H-PLM go down with the increasing of the sequence length.
Our hypothesis is that WebFormer utilizes L2L relative attention and the H2L attention, which enables the model to encode web documents with long sequences effectively and efficiently. Note that the DNN model does not depend on the sequence length and thus does not suffer from the long sequence.
\begin{table}
%\small
\begin{adjustbox}{width=0.95\columnwidth,center}
\begin{tabular}{c|c|c|c}
\hline
 & parameters & SWDE  & Common Crawl \\
\hline
AVEQA & 110M & 83.27 & 78.45  \\
H-PLM & 110M & 83.42 & 80.78  \\
\hline
WebFormer-2L & 45M & 82.05 & 76.73 \\
WebFormer-6L & 82M & 83.86 & 79.35 \\
WebFormer-12L-share & 109M & 85.29 & 81.49 \\
WebFormer-12L & 151M & 86.58 & 83.22 \\
WebFormer-24L & 285M & {\bf 87.84} & {\bf 86.51} \\
\hline
\end{tabular}
\end{adjustbox}
\caption{EM results over different model configurations.}\label{ablation}
\end{table}
\begin{table*}
\begin{center}
\begin{tabular}{c|ccc|ccc|ccc}
\hline
batch size & \multicolumn{3}{c|}{64} &\multicolumn{3}{c|}{128} &\multicolumn{3}{c}{512} \\
\hline
learning rate & $3$x$10^{-5}$ &   $5$x$10^{-5}$ & $1$x$10^{-4}$ & $3$x$10^{-5}$ &   $5$x$10^{-5}$ & $1$x$10^{-4}$ & $3$x$10^{-5}$ &   $5$x$10^{-5}$& $1$x$10^{-4}$ \\
\hline
SWDE  & {\bf 86.58} & 86.36 & 86.20 & 86.37& 86.42 & 86.35 & 86.18& 86.11 & 86.28\\
Events & {\bf 84.79}  & 84.62 & NaN & 84.54 & 84.46 & 84.65 & 84.11 & 84.27 & 84.13\\
Products & 80.67  & {\bf 80.71} & NaN &  80.32 & 80.38  & 80.40 &  79.96 & 80.23  & 80.37  \\
Movies & {\bf 85.30} & 85.21 & 85.14 & 84.58 & 84.75& 84.83 & 84.39 & 84.56 & 84.77 \\
\hline
\end{tabular}
\end{center}
\caption{EM results of WebFormer with different batch sizes and learning rates on all datasets.}\label{bs_lr}
\end{table*}

\subsubsection{Error Analysis}
We conduct error analysis of WebFormer over 160 and 60 randomly selected Exact Match mistakes on SWDE and Common Crawl dataset respectively (5 per field). We identify several major mistake patterns and summarize them here: 1) Partial text extraction: The largest group of mistakes is that our model extracts a substring of the ground-truth text. For example, our model extracts ``Fun Festival'' as the event name instead of ``Fun Festival at Square Park''.
2) Multiple occurrences issue: There are cases where the target field is mentioned multiple times on the web page. For example, our model extracts ``SEP 11'' as the date, but the ground-truth text is ``Sat, September 11, 2011''.
3) Multi-value issue: The other type of error is that the field has multiple values and we only extract one of them. For example, a product has both ``blue'' and ``white'' as its color where we only extract ``blue''.
4) Range issue: There are a certain amount of mistakes that fall into the range issue group. For instance, our model extracts the ``price'' as ``19.90'' from the ground-truth ``19.90 - 26.35'' which is a range of prices.
5) Model mistakes: There are few other extraction errors made by the model, which are hard cases even for human raters.
The summarization of the mistake analysis is reported in Figure \ref{fig:mistake}. By looking closely at these mistake patterns, we observe that our model actually extracts the correct or partially correct answers for most cases in the group of 1), 2), 3) and 4). These mistakes can be easily fixed by marking all answer occurrences and values as positives in the training, and adopting a BIO-based span extraction as mentioned. However, there are still difficult cases which require further investigations into the training data and the model.

\subsubsection{Ablation Study}
We further conduct a series of ablation studies of WebFormer. The WebFormer base model contains 12 layers. We first evaluate our model with a different number of encoder layers, i.e. 2L, 6L and 24L. We also evaluate another ablation of WebFormer by sharing the model parameters. Specifically, the query matrices of the text and HTML tokens are shared, i.e. $W_Q^{T2T}$=$W_Q^{T2H}$=$W_Q^T$, $W_Q^{H2H}$=$W_Q^{H2T}$=$W_Q^H$, $W_K^{T2T}$=$W_K^{H2T}$=$W_K^T$ and $W_K^{H2H}$=$W_K^{T2H}$=$W_K^H$. This model is referred to as WebFormer-12L-share. The EM results with the number of model parameters are shown in Table \ref{ablation}. It can be observed that WebFormer-24L achieves the best performance, which is consistent with our expectations. Similar behavior is also observed in ~\cite{DevlinCLT19,AinslieOACFPRSW20}. However, a larger model usually requires longer training time, as well as inference. The training time of the base models are reported in Table \ref{table:data}.

\subsubsection{Impact of Training Batch Size and Learning Rate}
To evaluate the model performance with different training batch size and learning rate,
we conduct experiments to train a set of WebFormer models with a hyper-parameter sweep consisting of learning rates in \{$3$x$10^{-5}$, $5$x$10^{-5}$, $1$x$10^{-4}$\} and batch-size in \{64, 128, 512\} on the training set.
The EM results with different learning rates and batch sizes on all datasets are reported in Table \ref{bs_lr}.
It can be seen from the tables that WebFormer achieves the best result with batch size 64 and learning rate $3$x$10^{-5}$ on all datasets except Products. The observation is consistent with the findings in work \cite{WangYKSSSYE20}, where smaller batch size usually leads to better performance.
This is also the reason that we set batch size to 64 and learning rate to $3$x$10^{-5}$ in all our previous experiments.
\begin{figure}
\centering
\includegraphics[width=0.95\linewidth]{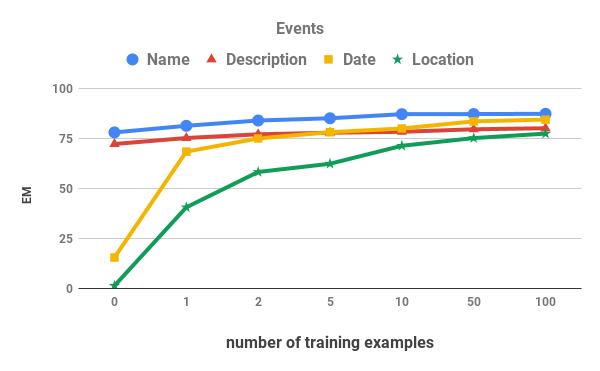}
\caption{EM results on zero-shot and few-shot learning.} \label{fig:generalization}
\end{figure}

\subsubsection{Zero-shot/Few-shot Extraction}
We conduct zero-shot and few-shot extraction experiments to evaluate the generalization ability of WebFormer on unseen domains/fields. In this experiment, we first pre-train a WebFormer model on Products and Movies data only. We then perform fine-tuning on Events data for 10K steps by varying the number of training examples from \{0, 1, 2, 5, 10, 50, 100\}. The EM scores of WebFormer on all four event fields are shown in Figure~\ref{fig:generalization}.
There are several interesting observations from this table. First, when the number of training examples is 0 (zero-shot extraction), the EM scores on ``Name'' and ``Description'' are reasonable around 75\%. However, the score on ``Location'' is close to 0. The reason is that both ``Name'' and ``Description'' are general fields that appear across domains, e.g. they both present in Products and Movies data. Therefore, the learned knowledge in WebFormer can be directly transferred to a new domain - Events. On the other hand, the pretrained model lacks knowledge about ``Location'' and thus performs poorly on this field. Second, it is not surprising to see that the EM scores increase with more training examples, and reach reasonably high values with 100 training examples. We also observe that the EM score for ``Location'' boosts dramatically even with one or two training examples.

\section{Conclusion}
%Structure information extraction from web pages is an important research topic that is widely studied in information retrieval and web search.
In this paper, we introduce a novel Web-page transFormer model, namely WebFormer, for structure information extraction from web documents. The structured HTML layout information is jointly encoded through the rich attention patterns with the text information. WebFormer effectively recovers both local syntactic and global layout information from web document serialization.
An extensive set of experimental results on SWDE and Common Crawl benchmarks has demonstrated the superior performance of the proposed approach over several state-of-the-art methods. In future, we plan to extend this work to multimodal learning that incorporates visual features. %Moreover, we also plan to conduct structured information extraction for web documents with multi-objects.

\begin{acks}
This work is supported by the National Natural Science Foundation of China (No. 62176270).
\end{acks}

\bibliographystyle{ACM-Reference-Format}
\bibliography{main}

\end{document}